\documentclass[letterpaper]{article} 
\usepackage{arxiv}  
\usepackage{times}  
\usepackage{helvet}  
\usepackage{courier}  
\usepackage[hyphens]{url}  
\usepackage{graphicx} 
\usepackage{natbib}  
\usepackage{caption} 
\usepackage{algorithm}
\usepackage{algorithmic}
\usepackage[colorlinks=true, linkcolor=black, citecolor=black, urlcolor=black]{hyperref}

\usepackage{newfloat}
\usepackage{listings}
\usepackage{subfigure}
\DeclareCaptionStyle{ruled}{labelfont=normalfont,labelsep=colon,strut=off} 
\floatstyle{ruled}
\newfloat{listing}{tb}{lst}{}
\floatname{listing}{Listing}
\title{Event-Enhanced Blurry Video Super-Resolution}
\author{
    Dachun Kai\textsuperscript{\rm 1},
    Yueyi Zhang\textsuperscript{\rm 1,2}\thanks{Corresponding author.},
    Jin Wang\textsuperscript{\rm 1},
    Zeyu Xiao\textsuperscript{\rm 3},
    Zhiwei Xiong\textsuperscript{\rm 1,2},
    Xiaoyan Sun\textsuperscript{\rm 1,2}
}
\affiliations{
    \textsuperscript{\rm 1}MoE Key Laboratory of Brain-inspired Intelligent Perception and Cognition,\\University of Science and Technology of China\\
    \textsuperscript{\rm 2}Institute of Artificial Intelligence, Hefei Comprehensive National Science Center\\
    \textsuperscript{\rm 3}National University of Singapore\\
    \{dachunkai, jin01wang\}\noindent@mail.ustc.edu.cn, zeyuxiao\noindent@nus.edu.sg, \{zhyuey, zwxiong, sunxiaoyan\}\noindent@ustc.edu.cn
}

\usepackage{bibentry}

\usepackage{amssymb}
\usepackage{amsmath}
\usepackage{multirow}
\usepackage{xcolor}
\usepackage{booktabs}
\usepackage{colortbl}
\usepackage{makecell}

\begin{document}

\maketitle

\begin{abstract}
In this paper, we tackle the task of blurry video super-resolution (BVSR), aiming to generate high-resolution (HR) videos from low-resolution (LR) and blurry inputs. Current BVSR methods often fail to restore sharp details at high resolutions, resulting in noticeable artifacts and jitter due to insufficient motion information for deconvolution and the lack of high-frequency details in LR frames. To address these challenges, we introduce event signals into BVSR and propose a novel event-enhanced network, Ev-DeblurVSR. To effectively fuse information from frames and events for feature deblurring, we introduce a reciprocal feature deblurring module that leverages motion information from intra-frame events to deblur frame features while reciprocally using global scene context from the frames to enhance event features. Furthermore, to enhance temporal consistency, we propose a hybrid deformable alignment module that fully exploits the complementary motion information from inter-frame events and optical flow to improve motion estimation in the deformable alignment process. Extensive evaluations demonstrate that Ev-DeblurVSR establishes a new state-of-the-art performance on both synthetic and real-world datasets. Notably, on real data, our method is +2.59dB more accurate and 7.28$\times$ faster than the recent best BVSR baseline FMA-Net. 
\end{abstract}

%
\begin{links}
    \link{Code}{https://github.com/DachunKai/Ev-DeblurVSR}
\end{links}

\section{Introduction}

Video super-resolution (VSR) aims to recover a high-resolution (HR) video from its low-resolution (LR) counterpart. While existing methods~\cite{xu2024enhancing,zhou2024video} get good results for general videos, they struggle with hard cases involving severe motion blur. Yet, such a setting is very common in practical VSR applications, like sports broadcasting~\cite{liu2021large} and video surveillance~\cite{shamsolmoali2019deep}. For example, in sports videos, fast-moving objects often cause unwanted motion blur.

\begin{figure}[t!]
	\centering
	\includegraphics[width=\columnwidth]{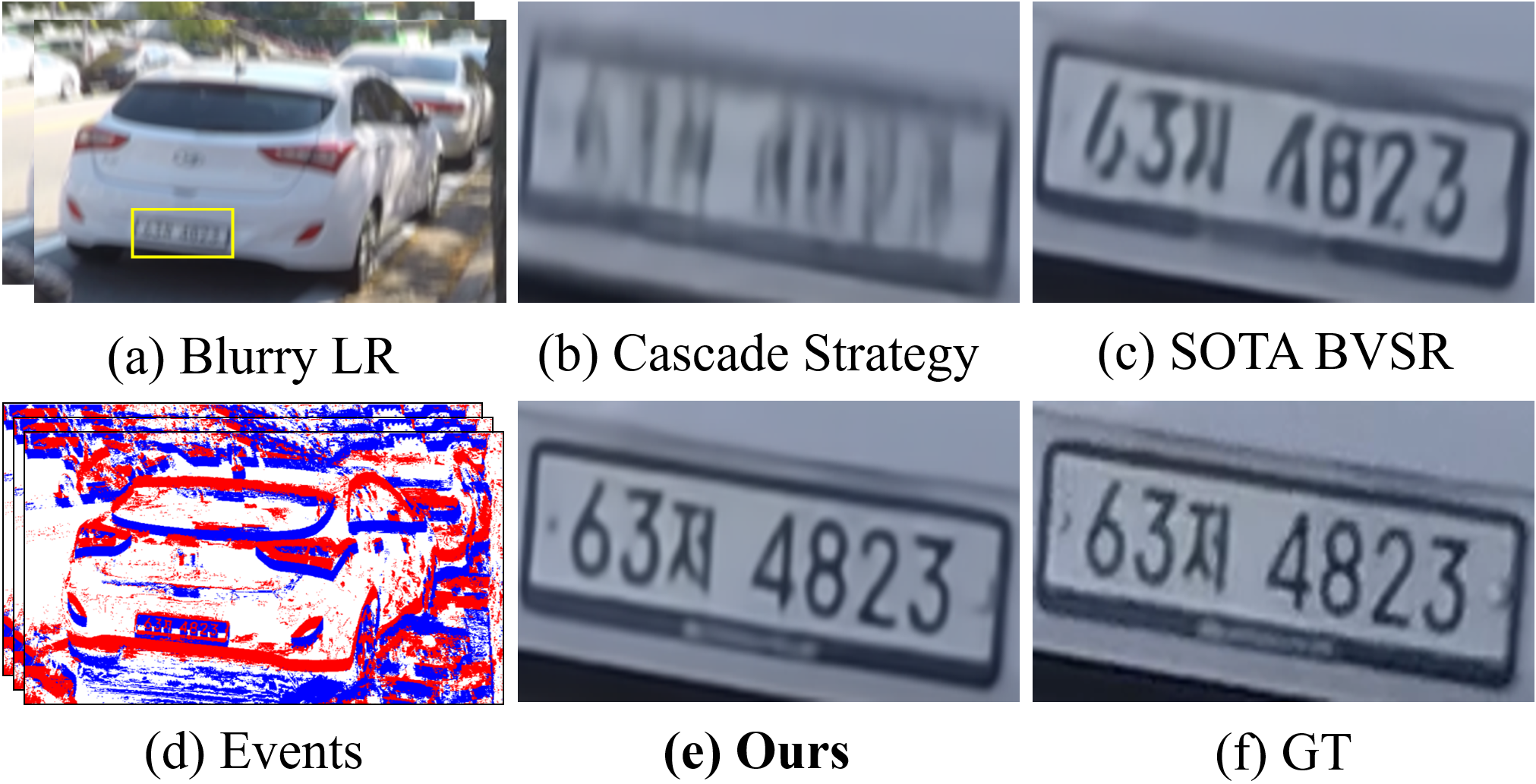}
        \caption{An example (a) from a challenging blurry video enhanced by (b) SOTA methods in video deblurring~\cite{zhang2024blur} + VSR~\cite{xu2024enhancing}; (c) a SOTA BVSR method~\cite{youk2024fma}; and (e) our event-enhanced approach. Our method can restore the license plate with finer details, sharper edges, and less aliasing.}
	\label{fig:fig1}
        \vspace{-1ex}
\end{figure}

To achieve VSR from a blurry video, \textit{i.e.}, blurry VSR (BVSR), a straightforward approach is to perform video deblurring, followed by VSR methods, which we refer to as the \textit{cascade} strategy. However, this approach has a drawback in that the pixel errors introduced in the deblurring stage are propagated and amplified in the subsequent VSR step, thus degrading the overall performance. To address this, some works~\cite{fang2022high,youk2024fma} have proposed joint learning methods of VSR and deblurring. For instance, FMA-Net~\cite{youk2024fma} proposes jointly estimating the degradation and restoration kernels through sophisticated representation learning. However, as shown in Fig.~\ref{fig:fig1}, these methods still suffer from blurry artifacts, jitter effects, and temporal aliasing.

Relying solely on blurry LR frames to restore high-quality HR videos is a highly ill-posed problem. This is due to the inherent lack of motion information needed to deconvolve blurred images and the lack of high-frequency details in LR frames. Recently, event signals captured by event cameras have been used for image deblurring~\cite{xu2021motion,yang2023event}. Compared to standard cameras, event cameras have very high temporal resolution, high dynamic range~\cite{gallego2020event}, and rich ``moving edge" information~\cite{mitrokhin2020learning}. These characteristics enable events to provide complementary motion information, as well as high-frequency details, for BVSR. Motivated by these advantages, we propose including event signals as auxiliary information to enhance BVSR performance.

In this paper, we present Ev-DeblurVSR, a novel event-enhanced network for BVSR. To effectively fuse information from frames and events, we first categorize events into intra-frame and inter-frame events. Intra-frame events provide valuable motion and high-frequency information during the frames' exposure time, aiding in deblurring frame features. Frames, in turn, offer global scene context, further enhancing event features. This synergy motivates our Reciprocal Feature Deblurring (RFD) module. Inter-frame events capture continuous motion between frames, which is crucial for temporal consistency. For this purpose, we propose a Hybrid Deformable Alignment (HDA) module that combines inter-frame event information with optical flow for superior motion estimation in the deformable alignment process. Experimental results on three datasets demonstrate the effectiveness of our proposed Ev-DeblurVSR. Our Ev-DeblurVSR significantly outperforms existing methods in both spatial recovery and temporal consistency. To summarize, our main contributions are:

\begin{itemize}
    \item We present Ev-DeblurVSR, the \textbf{first} event-enhanced scheme for BVSR. Our Ev-DeblurVSR leverages motion information and high-frequency details from both intra-frame and inter-frame events for BVSR.
    \item We propose the RFD module, which effectively utilizes mutual assistance between frames and intra-frame events to facilitate feature deblurring.
    \item We propose the HDA module, which fully exploits complementary motion information from optical flow and inter-frame events to improve temporal alignment.
    \item Ev-DeblurVSR achieves state-of-the-art performance on three datasets, including synthetic and real-world data.
\end{itemize}

\section{Related Work}

\paragraph{Video Super-Resolution.} With the rapid development of deep learning, there has been significant progress in VSR~\cite{wang2019edvr,xiao2020space,xiao2021space,liu2022learning,xia2023structured,li2024savsr}. Compared to single-image super-resolution, VSR focuses more on modeling temporal relationships and aligning frames. For example, BasicVSR++~\cite{chan2022basicvsr++} introduced second-order grid propagation and flow-guided deformable alignment to explore long-term information across misaligned frames. However, these methods often perform poorly in challenging cases, such as videos with severe motion blur~\cite{li2024object}. To address this issue, ~\citet{fang2022high} proposed the first deep learning-based BVSR network that uses a parallel-fusion module to combine features from SR and deblurring branches. Recently, ~\citet{youk2024fma} presented FMA-Net, a method for joint learning of spatiotemporally variant degradation and restoration kernels through complex motion representation learning. However, these methods often fail when there are large pixel displacements, resulting in severe temporal inconsistency.

\paragraph{Video Deblurring.} Video deblurring aims to recover sharp videos from blurry inputs, where exploring temporal information is crucial~\cite{li2021arvo,jiang2022erdn,zhu2022deep,cao2022vdtr,lin2022flow,wang2023unsupervised,li2024object,liang2024vrt}. To efficiently transfer useful information from neighboring frames, \citet{zhong2020efficient} proposed using a global spatiotemporal attention module within a recurrent framework to propagate information from non-local frames. However, incorrect estimation of non-local frames can lead to error propagation through the recurrent process. To address this, \citet{pan2023deep} introduced a deep discriminative spatial and temporal network with a channel-wise gated dynamic module to adaptively explore useful information from non-local frames for better video restoration. More recently, \citet{zhang2024blur} proposed a Blur-aware Spatio-Temporal Sparse Transformer Network (BSSTNet) for video deblurring. BSSTNet uses a blur map to convert dense attention into a sparse form, allowing for more extensive information utilization throughout the entire video sequence. This approach has shown significant performance improvements in the area of video deblurring.

\begin{figure}[t!]
	\centering
	\includegraphics[width=\columnwidth]{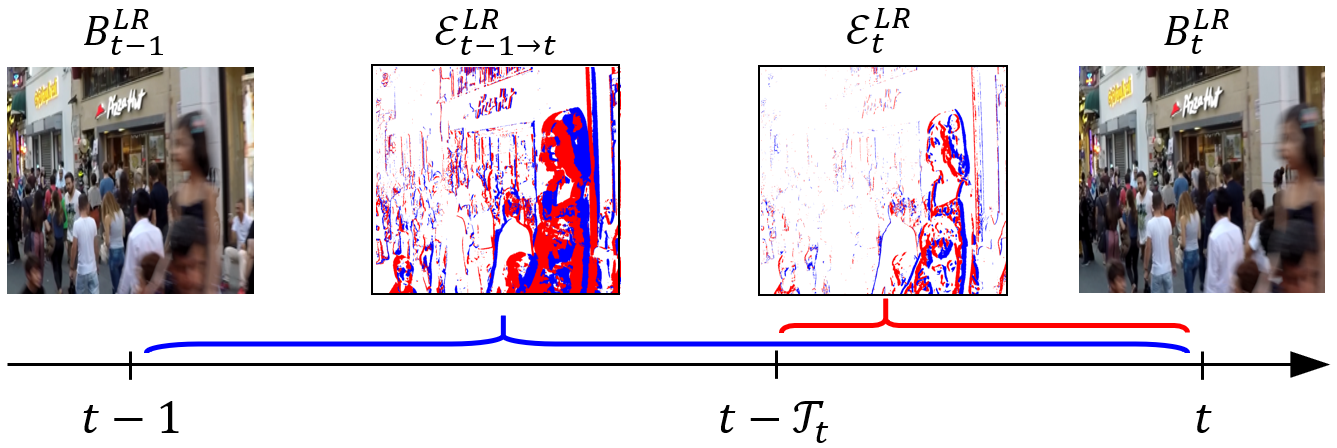}
	\caption{Proposed event processing for BVSR. The exposure time of $B^{LR}_t$ is $\mathcal{T}_t$. Intra-frame events $\mathcal{E}^{LR}_{t}$ capture motion within the exposure, which is used to deblur frame features. Inter-frame events $\mathcal{E}^{LR}_{t-1\to t}$ capture motion between frames, which helps enhance temporal alignment in VSR.}
	\label{fig:fig2}
        \vspace{-1ex}
\end{figure}

\paragraph{Event-based Vision.} Event cameras, also known as dynamic vision sensors~\cite{lichtsteiner2008128}, are new bio-inspired vision sensors that measure pixel-wise brightness changes asynchronously and output events. They offer super high temporal resolution (about $1\mu$s) and high dynamic range (140 dB)~\cite{gallego2020event,chen2021indoor}. With these advantages, event signals have been widely applied in optical flow estimation~\cite{shiba2022secrets,luo2024efficient} and video frame interpolation~\cite{xiao2022eva,kim2023event,liu2024video}. With their high temporal resolution, event data can provide rich motion information~\cite{xiao2024estme,xiao2024micro} during the frame's exposure time, which helps deconvolve blurred images~\cite{yang2022learning,yang2024learning,yang2024motion,yang2024latency,kim2024frequency,yu2024learning}.~\citet{sun2022event} devised an event-image fusion module to adaptively integrate event features with image features, alongside a symmetric cumulative event voxel representation for event-based frame deblurring. 

\begin{figure*}[t!]
	\centering
	\includegraphics[width=\textwidth]{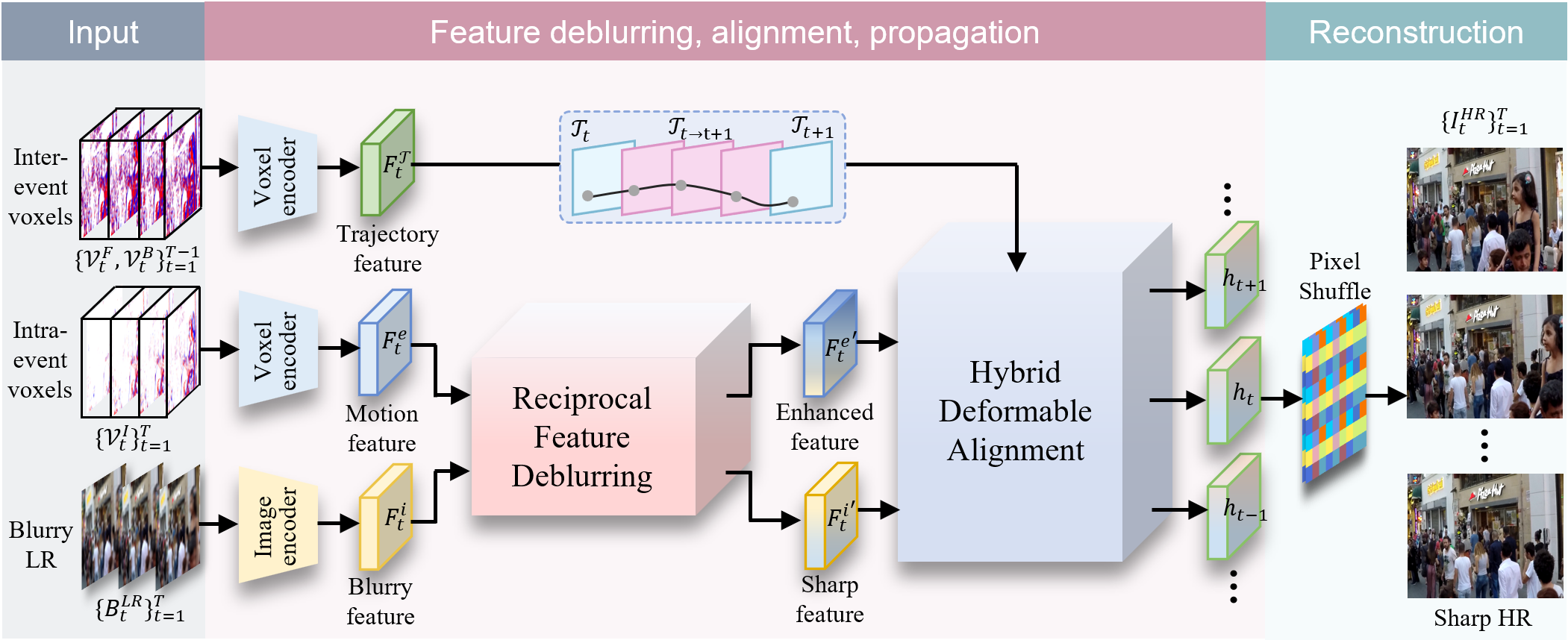}
	\caption{Overview of Ev-DeblurVSR. Intra-frame voxels are fused with blurry frames in the RFD module to deblur frame features and enhance event features with scene context. Inter-frame voxels are integrated into the HDA module, using continuous motion trajectories to guide deformable alignment. Finally, the aligned features are upsampled to reconstruct sharp HR frames.}
	\label{fig:fig3}
        \vspace{-1ex}
\end{figure*}

Recent studies~\cite{jing2021turning,kai2023video,lu2023learning,kai2024evtexture,xiao2024event,xiao2024asymmetric} have proposed combining an event camera with an RGB camera to improve VSR performance. Typically,~\citet{jing2021turning} proposed E-VSR, which first utilizes events to reconstruct intermediate frames. The high-frame-rate video is then encoded into a VSR module to recover HR videos. However, these methods generally assume that the input frames are sharp. In VSR, training with events and blurry frames remains a challenging problem.

\section{Method}

\subsection{Event Processing for BVSR}
\label{subsec:3_1}

Previous event-based VSR methods are insufficient for BVSR as they only use inter-frame events for alignment. However, temporal misalignment between the timestamps of blurry frames and the inter-frame events hinders the modeling of inherent motion within the blurry frames.

To address this, we propose categorizing events into intra-frame and inter-frame events for BVSR. As shown in Fig.~\ref{fig:fig2}, given two blurry LR frames, $B_{t-1}^{LR}$ and $B_{t}^{LR}$, and the event stream $\mathcal{E}$, with $B_{t}^{LR}$ having an exposure time of $\mathcal{T}_t$, our goal is to deblur and upscale these frames to sharp HR frames, $I_{t-1}^{HR}$ and $I_{t}^{HR}$. Intra-frame events $\mathcal{E}_{t}$ capture motion information within the exposure time of each blurry frame and are used for deblurring frame features. Inter-frame events $\mathcal{E}_{t-1\to t}$ capture continuous motion trajectories between frames and are used for feature alignment in VSR. 

We represent events as a grid-like event voxel grid $\mathcal{V}$ as in~\cite{zhu2019unsupervised}. In our experiments, we set the number of bins to 5, consistent with the earlier study~\cite{weng2021event}. We can thus obtain intra-frame voxels $\mathcal{V}^{I}_t$. For inter-frame events, since our model uses a bidirectional recurrent network as in BasicVSR~\cite{chan2021basicvsr}, we generate forward voxels $\mathcal{V}^{F}_t$ and backward voxels $\mathcal{V}^{B}_t$.

\subsection{Ev-DeblurVSR Network}
\label{subsec:3_2}

\subsubsection{Framework Overview.}

The architecture of our proposed Ev-DeblurVSR is shown in Fig.~\ref{fig:fig3}. The network's input includes a blurry LR sequence consisting of $T$ frames, denoted as $\{B^{LR}_t\}_{t=1}^{T}$, along with their intra-frame voxels $\{\mathcal{V}^{I}_t\}_{t=1}^{T}$, and the $T-1$ intervals' inter-frame voxels, including forward voxels $\{\mathcal{V}^{F}_t\}_{t=1}^{T-1}$ and backward voxels $\{\mathcal{V}^{B}_t\}_{t=1}^{T-1}$. The output is a sharp HR sequence $\{I^{HR}_t\}_{t=1}^{T}$.

The proposed Ev-DeblurVSR comprises two key components: the RFD module and the HDA module. Firstly, the input voxels and frames are passed through their respective feature extractors, comprising five residual blocks as used in~\cite{wang2018esrgan}, yielding trajectory, motion, and blurry frame features.
In the RFD module, we leverage motion information from intra-frame event features to deblur frame features. Reciprocally, we also enhance event features with global scene context from frame features. In the HDA module, we utilize motion trajectory information from inter-frame events and optical flow to collaboratively enhance motion estimation for the deformable alignment process in VSR. Finally, the aligned features are processed through pixel shuffle~\cite{shi2016real} layers and added with bicubic upsampled results to reconstruct sharp HR frames.  

\subsubsection{Reciprocal Feature Deblurring.}
\label{cfd}

\begin{figure}[t!]
	\centering
	\includegraphics[width=\columnwidth]{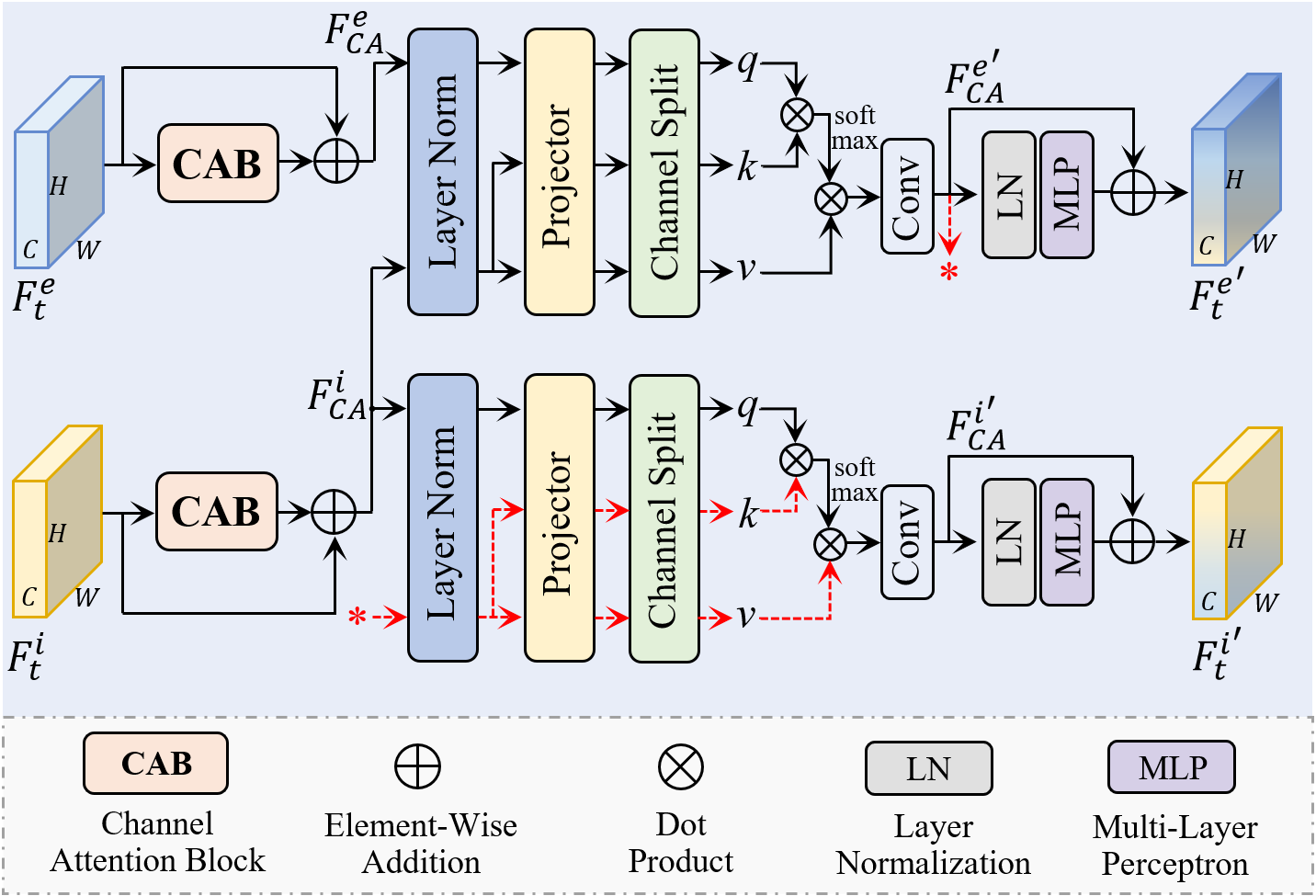}
	\caption{The structure of the RFD module.}
	\label{fig:fig4}
        \vspace{-1ex}
\end{figure}

To address the limitations of events in feature deblurring due to sparsity and limited scene context~\cite{messikommer2020event}, we propose the RFD module. This module not only utilizes events for effective deblurring but also integrates frames to enhance event features. As shown in Fig.~\ref{fig:fig4}, at timestamp $t$, the RFD module receives the blurry frame feature $F_t^i$ and the intra-frame event feature $F_t^e$ as inputs. They are processed through two pathways, the event and frame pathways, each including a multi-head Channel Attention Block (CAB). The frame pathway captures global scene context, producing $F_{CA}^i$, while the event pathway learns motion information, resulting in $F_{CA}^e$. The operation is as follows:
\begin{equation}
    F_{CA}^e = \textbf{CAB}(F_t^e),\ F_{CA}^i = \textbf{CAB}(F_t^i).
\end{equation}
The frame feature $F_{CA}^i$ is then fed into the event pathway to enhance event features with scene details, resulting in $F_{CA}^{e'}$. This output is then used to deblur $F_{CA}^i$, producing $F_{CA}^{i'}$. The above operations are performed using a QKV-based multi-head cross-modal attention mechanism as follows:
\begin{align}
    F_{CA}^{e'} &= F_{CA}^e + \mathbf{V}_{i} \operatorname{softmax}\left(\tfrac{\mathbf{Q}_{e}^T \mathbf{K}_{i}}{\sqrt{C}}\right), \\
    F_{CA}^{i'} &= F_{CA}^i + \mathbf{V}_{e} \operatorname{softmax}\left(\tfrac{\mathbf{Q}_{i}^T \mathbf{K}_{e}}{\sqrt{C}}\right),
\end{align}
where we use a $1 \times 1$ convolutional layer to create attention maps. After that, we apply layer normalization and MLP layers to aggregate information. This results in scene-enhanced event feature $F^{e'}_t$ and sharper frame feature $F^{i'}_t$.

\subsubsection{Hybrid Deformable Alignment.}
Events and optical flow can both represent motion, but they have different characteristics~\cite{wan2022learning}. Events provide continuous motion but are spatially sparse. Optical flow offers rich spatial information but lacks temporal continuity. To leverage this complementarity, we propose integrating optical flow and events to improve motion estimation in deformable alignment used in VSR~\cite{shi2022rethinking}.

\begin{figure}[t!]
	\centering
	\includegraphics[width=\columnwidth]{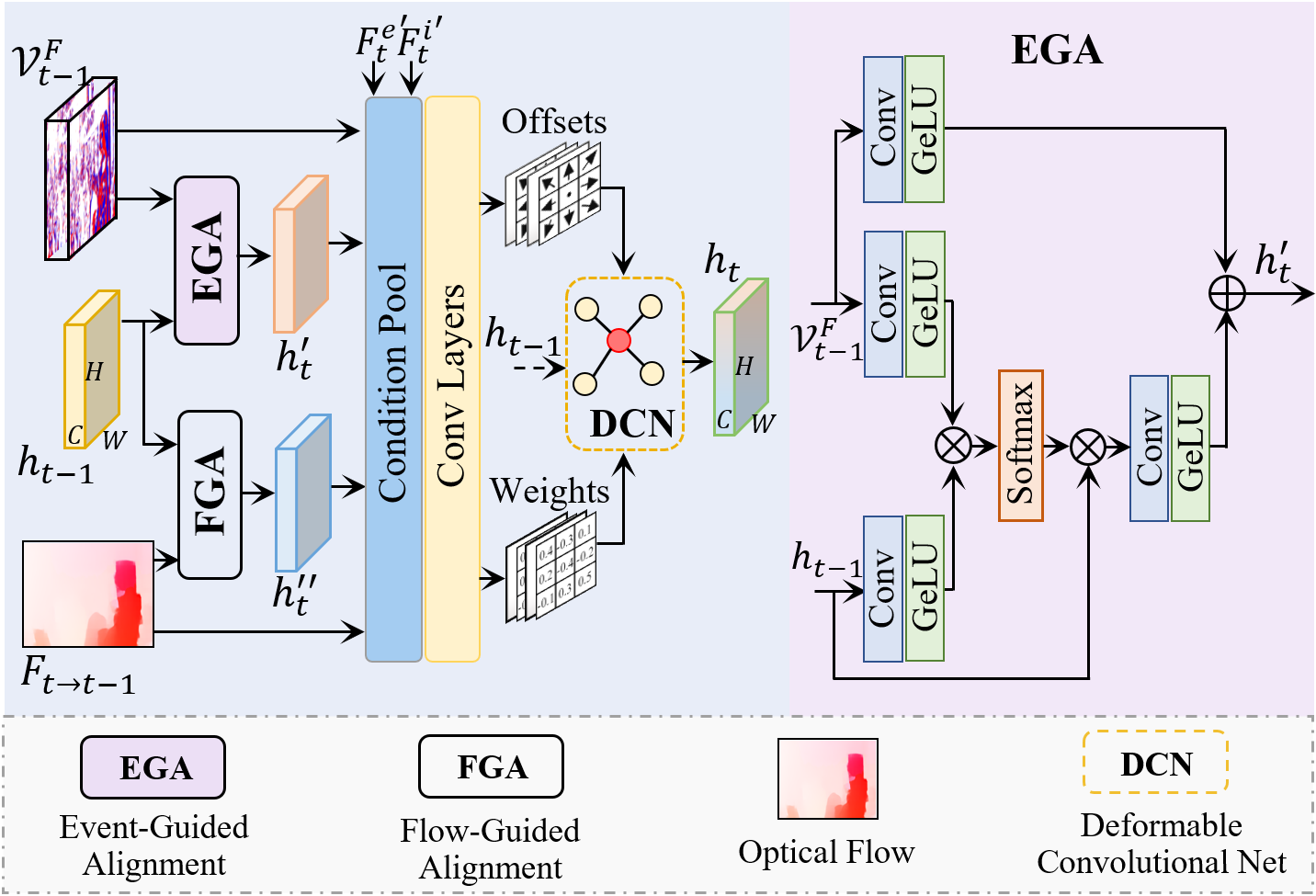}
	\caption{The structure of the HDA module.}
	\label{fig:fig5}
        \vspace{-1ex}
\end{figure}

We introduce the HDA module, with its structure shown in Fig.~\ref{fig:fig5}. We use the feature propagation process from $t-1$ to $t$ as an example to illustrate the alignment process. The HDA module adopts a two-branch structure: the Event-Guided Alignment (EGA) branch and the Flow-Guided Alignment (FGA) branch. In the EGA branch, we use the inter-frame voxel $\mathcal{V}^F_{t-1}$ to align $h_{t-1}$ to $h_t'$. The FGA branch employs well-established SpyNet~\cite{ranjan2017optical} to estimate optical flow $F_{t \to t-1}$. This flow is then used to backward warp $h_{t-1}$, generating the flow-based alignment feature $h''$. The process is as follows:
\begin{equation}
    h_t' = \textbf{EGA}(h_{t-1},\ \mathcal{V}^F_{t-1}),\ h_t'' = \textbf{FGA}(h_{t-1},\ F_{t \to t-1}).
\end{equation}

In our EGA, we first apply a convolutional layer to $\mathcal{V}^F_{t-1}$ and $h_{t-1}$ to match their channel dimensions. They are then element-wise multiplied, followed by a softmax operation to compute channel-wise similarity scores. The similarity information is used to modulate $h_{t-1}$, which incorporates the event information into the alignment process. The modulated feature is then combined with the processed $\mathcal{V}^F_{t-1}$ to produce the event-based alignment feature $h'$. 
\begin{equation}\label{eq5*}
    h_t = \textbf{DCN}(h_{t-1}; h_t', h_t'', F^{e'}_t, F^{i'}_t, F_{t \to t-1}).
\end{equation}

Finally, $h_t'$ and $h_t''$, along with $F_{t \to t-1}$, are concatenated with $F^{e'}_t$ and $F^{i'}_t$. As in Eq.~(\ref{eq5*}), these features form our condition pool and are fed into a stack of convolutional layers to predict the motion offsets and modulation weights for the Deformable Convolutional Network (DCN). The learned offsets and weights are used to deform and align $h_{t-1}$ to $h_t$.

\subsection{Loss function}
\label{subsec:3_3}

Previous VSR studies~\cite{chan2021basicvsr,chan2022basicvsr++} typically use MSE loss, denoted as $\mathcal{L}_r$, for supervision, calculated between the ground-truth (GT) and super-resolved HR clips. However, this loss treats all pixels equally, regardless of high-frequency and low-frequency regions. It also averages the errors of all pixels, leading to over-smooth results~\cite{xie2023mitigating}. To address this, we propose an edge-enhanced loss $\mathcal{L}_e$ that utilizes high-frequency event information to selectively weight pixel reconstruction errors:
\begin{equation}\label{eq5}
	\mathcal{L}_e=\frac{1}{T} \sum_{t=1}^T \lvert \mathcal{V}_t^{HR} \rvert 
	\cdot \sqrt{\left\|I^{GT}_t-I^{H R}_t\right\|^2+\eta^2}.
\end{equation}
Here, $\mathcal{V}_t^{HR} \in \mathbb{R}^{sH \times sW \times 3}$ represents the edge-related mask derived from HR voxels within exposure $\mathcal{T}_t$, ranging from $[-1.0, +1.0]$, where $s$ is the upsampling factor. And $\eta$ is a small smoothing factor to avoid numerical instability. In our experiments, we set $\eta=1\times10^{-8}$. Our final loss function is a combination of \( \mathcal{L}_r \) and \( \mathcal{L}_e \), \textit{i.e.}, \( \mathcal{L} = \mathcal{L}_r + \mathcal{L}_e \).

\begin{table*}[t!]
	\centering
	\resizebox{0.98\textwidth}{!}{  
		\begin{tabular}{ccccccccc}
			\toprule
			\multicolumn{2}{c}{\multirow{2}[1]{*}{Method Type}} & \multirow{2}[1]{*}{Method} & \multicolumn{3}{c}{GoPro} & \multirow{2}[1]{*}{\makecell{\#Params \\ (M)}} & \multirow{2}[1]{*}{\makecell{FLOPs \\ (G / frame)}} & \multirow{2}[1]{*}{\makecell{Runtime \\ (ms / frame)}} \\
			& & & PSNR$\uparrow$ & SSIM$\uparrow$ & LPIPS$\downarrow$ & & & \\
			\midrule
			\multirow{8}[1]{*}{RGB-based} & \multirow{4}[1]{*}{\makecell{Deblur\\ +\\ VSR}} & DSTNet + BasicVSR++ & 24.43 & 0.7471 & 0.3816 & 7.45 + 7.32 & 44.9 + 405.6 & 7.0 + 64.4 \\
			& & DSTNet + IART & 24.43 & 0.7467 & 0.3842 & 7.45 + 13.41 & 44.9 + 1972.7 & 7.0 + 1321.2 \\
			& & BSSTNet + MIA-VSR & 26.40 & 0.8192 & 0.3161 & 48.18 + 16.60  & 314.7 + 1267.5   & 67.8 + 831.0   \\
			& & BSSTNet + IART & 26.40 & 0.8189 & 0.3148 & 48.18 + 13.41 & 314.7 + 1972.7 & 67.8 + 1321.2 \\
			\cmidrule{2-9}
			& \multirow{4}[1]{*}{\makecell{BVSR}} & BasicVSR++$^*$ & 30.79 & \underline{0.9077} & \underline{0.2287} & 7.32 & 405.6 & 64.4 \\
			& & MIA-VSR$^*$ & 27.91  & 0.8481  &  0.2901  & 16.60  & 1267.5  & 831.0  \\
			& & IART$^*$ & 27.69 & 0.8372 & 0.3050 & 13.41 & 1972.7 & 1321.2 \\
			& & FMA-Net & 29.24 & 0.8720 & 0.2682 & 9.62 & 1365.0 & 579.8 \\
			\midrule
			\multirow{9}[1]{*}{Event-based} & \multirow{4}[1]{*}{\makecell{Deblur\\ +\\ VSR}} & EFNet + EGVSR & 23.53 & 0.7276 & 0.4155 & 8.47 + 2.58 & 94.9 + 159.6 & 11.7 + 118.1 \\
			& & EFNet$^\dagger$ + EGVSR & 23.80 & 0.7422 & 0.3963 & 9.91 + 2.58 & 114.5 + 159.6 & 15.4 + 118.1 \\
			& & REFID + EvTexture & 23.72 & 0.7448 & 0.4019 &   15.92 + 8.90 & 89.1 + 805.4 & 16.2 + 100.8 \\
			& & REFID$^\dagger$ + EvTexture & 24.28 & 0.7738 & 0.3402 & 17.36 + 8.90 & 108.7 + 805.4 & 19.9 + 100.8 \\
			\cmidrule{2-9}
			& \multirow{5}[1]{*}{\makecell{BVSR}} & eSL-Net++ & 26.29 & 0.7959 & 0.3377 & \textbf{1.41} & 434.4 & \textbf{59.4} \\
			& & eSL-Net++$^\dagger$ & 26.43 & 0.8293 & 0.3052 & 2.85 & 454.0 & 63.1 \\
			& & EGVSR$^*$ & 27.79 & 0.8331 & 0.3037 & 2.58 & \textbf{159.6} & 118.1 \\
			& & EvTexture$^*$ & \underline{31.00} & 0.9065 & 0.2355 & 8.90 & 805.4 & 100.8 \\
			& & \textbf{Ev-DeblurVSR} & \textbf{32.51} & \textbf{0.9314} & \textbf{0.2041} & 8.28 & 459.5 & 79.6 \\
			
			\bottomrule
		\end{tabular}
	}
	\caption{Quantitative comparison on GoPro for $4\times$ BVSR. \textbf{All methods are retrained on the same dataset.} All results are calculated on the RGB channel. \textbf{Bold} and \underline{underlined} numbers represent the best and second-best performance. FLOPs and runtime are computed on one $180\times320$ LR frame. \textbf{$^*$ denotes the model initially proposed for sharp VSR, and we retrain it on blurry LR inputs. $^\dagger$ indicates the single-image model, and we include optical flow from SpyNet to refine it.}} 
	\label{tab:table1}
  \vspace{-1ex}
\end{table*}

\begin{table}[t!]
	\centering
	\resizebox{0.98\columnwidth}{!}{  
		\begin{tabular}{ccc}
			\toprule
			\multirow{2}[1]{*}{Method} & \multirow{2}[1]{*}{\makecell{BSD \\ PSNR/ SSIM/ LPIPS}} & \multirow{2}[1]{*}{\makecell{NCER \\ PSNR/ SSIM/ LPIPS}} \\ [1.2em]
			\midrule
			BasicVSR++$^*$ & \underline{31.12} / \underline{.9050} / \underline{.2580}	& 27.05 / \underline{.8255} /\underline{ .1975} \\
			MIA-VSR$^*$	& 29.24 / .8643  / .3074   &	24.55 / .7307 / .3251  \\
			IART$^*$ &	29.47 / .8689 / .2977 &	25.16 / .7499 / .2908 \\
			FMA-Net &	30.14 / .8805 / .2887 &	26.01 / .7779 / .2538 \\
			EGVSR$^*$ &	29.32 / .8665 / .3145 &	24.26 / .7218 / .3276 \\
			EvTexture$^*$ &	31.06 / .8956 / .2746 & \underline{27.23} / .8136 / .2241 \\
			\textbf{Ev-DeblurVSR} & \textbf{33.02} / \textbf{.9304} / \textbf{.2281} & \textbf{28.60} / \textbf{.8516} / \textbf{.1712} \\
			\bottomrule
		\end{tabular}
	}
	\caption{Comparison on BSD and NCER for $4\times$ BVSR.} 
	\label{tab:table2}
        \vspace{-1ex}
\end{table}

\section{Experiments}
\subsection{Datasets}

\paragraph{Synthetic datasets.} We use two widely-used datasets for training: \textbf{GoPro}~\cite{nah2017deep} and \textbf{BSD}~\cite{zhong2020efficient}. Then, we follow the strategy used in previous VSR studies by applying bicubic downsampling to the videos in the datasets to create blurry LR and sharp HR pairs. The GoPro dataset was recorded using a GoPro camera at 240 fps with a resolution of $1280\times720$. It contains 22 videos for training and 11 for testing. The blurry frames in this dataset are created by averaging several sharp frames. The BSD dataset, on the other hand, consists of real blurry-sharp video pairs captured using a beam splitter system. These videos have a resolution of $640\times480$ and a frame rate of 15 fps, which contain severe motion blur. The dataset includes 60 sequences for training and 20 for testing. Since the GoPro and BSD datasets do not have real event data, we use the commonly used event simulator Vid2E~\cite{gehrig2020video} to generate event data from the video clips.

\noindent\textbf{Real-world datasets.} 
We also train and test our method on real-world event data. For this, we use the recently published event-based motion deblurring dataset \textbf{NCER}~\cite{cho2023non}, which includes 27 videos for training (a total of 2,583 frames) and 16 videos for testing (1,454 frames). The dataset is recorded with a high-frame-rate (522 fps) RGB camera and a $640\times480$ DVXplorer event camera, covering various scenes and textures suitable for BVSR. 

\subsection{Implementation Details}

We follow the previous study~\cite{chan2022basicvsr++}; when training, we use 15 frames as inputs, set the mini-batch size to 8, and center-crop the input frames size and event voxels size as $64\times64$. We use random horizontal and vertical flips to augment the data. On the three datasets mentioned above, we first train the model on GoPro for 300K iterations using the Adam optimizer and Cosine Annealing scheduler. For the experiments on BSD, we fine-tune the model trained on GoPro with an initial learning rate of $1 \times 10^{-4}$ for 200K iterations. Then, similar to NCER, we fine-tune the model trained on BSD with the same hyperparameter settings. The entire training process runs on 8 NVIDIA RTX4090 GPUs and takes about four days per dataset to converge.

\begin{table}[t!]
	\centering
	\resizebox{\columnwidth}{!}{  
		\begin{tabular}{cccccc}
			\toprule
			GoPro & IART$^*$ & FMA-Net & EGVSR$^*$ & EvTexture$^*$ & \textbf{Ours} \\
                \midrule
			tOF$\downarrow$ & 2.94 & 2.30 & 2.78 & 1.73 & \textbf{1.43} \\
			TCC$\uparrow$\scalebox{0.5}{$ \times  10$} & 3.73 & 4.36 & 3.70 & 4.90 & \textbf{5.38} \\
			\midrule \midrule
			NCER & IART$^*$ & FMA-Net & EGVSR$^*$ & EvTexture$^*$ & \textbf{Ours} \\
                \midrule
			tOF$\downarrow$ & 1.61 & 0.96 & 1.39 & 0.72 & \textbf{0.54} \\
			TCC$\uparrow$\scalebox{0.5}{$ \times  10$} & 2.85 & 3.50 & 2.62 & 3.96 & \textbf{4.73} \\
			\bottomrule
		\end{tabular}
	}
	\caption{Temporal consistency on GoPro and NCER.} 
	\label{tab:table3}
  \vspace{-1ex}
\end{table}

 \begin{figure*}[t!]
    \centering
    \includegraphics[width=0.93\textwidth]{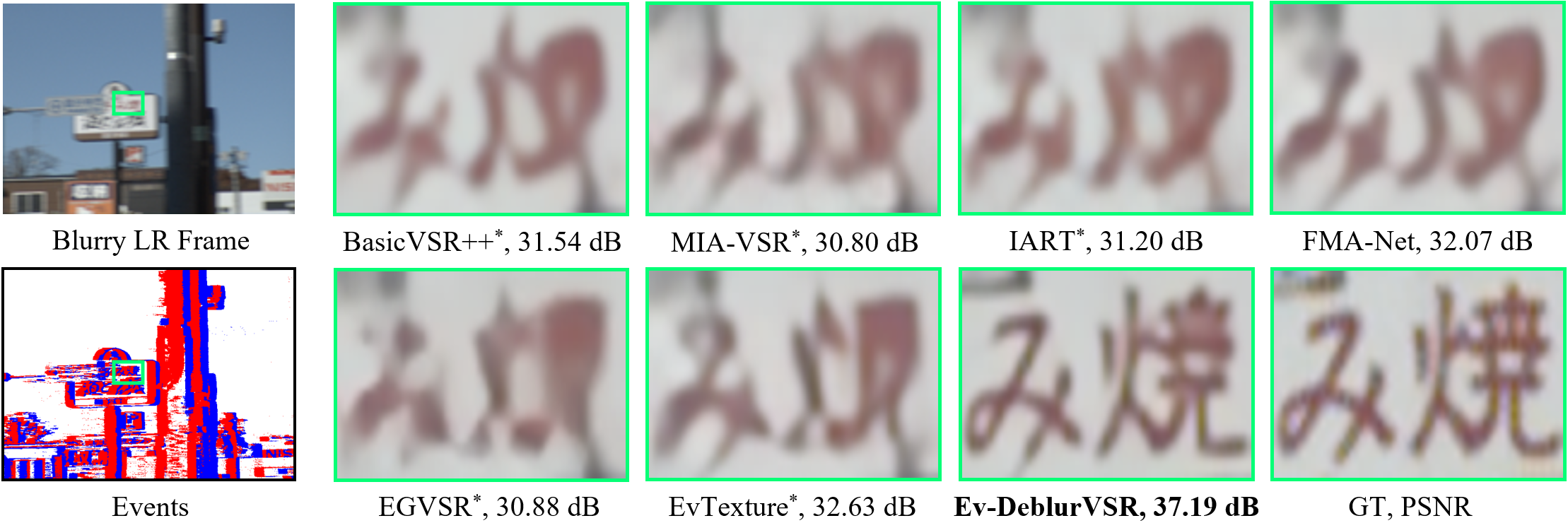}
    \caption{Qualitative comparison on BSD. Our method can restore clear road signs and text with sharp edges.}
    \label{fig:fig6}
    \vspace{-1ex}
\end{figure*}

\begin{figure*}[t!]
    \centering
    \includegraphics[width=0.93\textwidth]{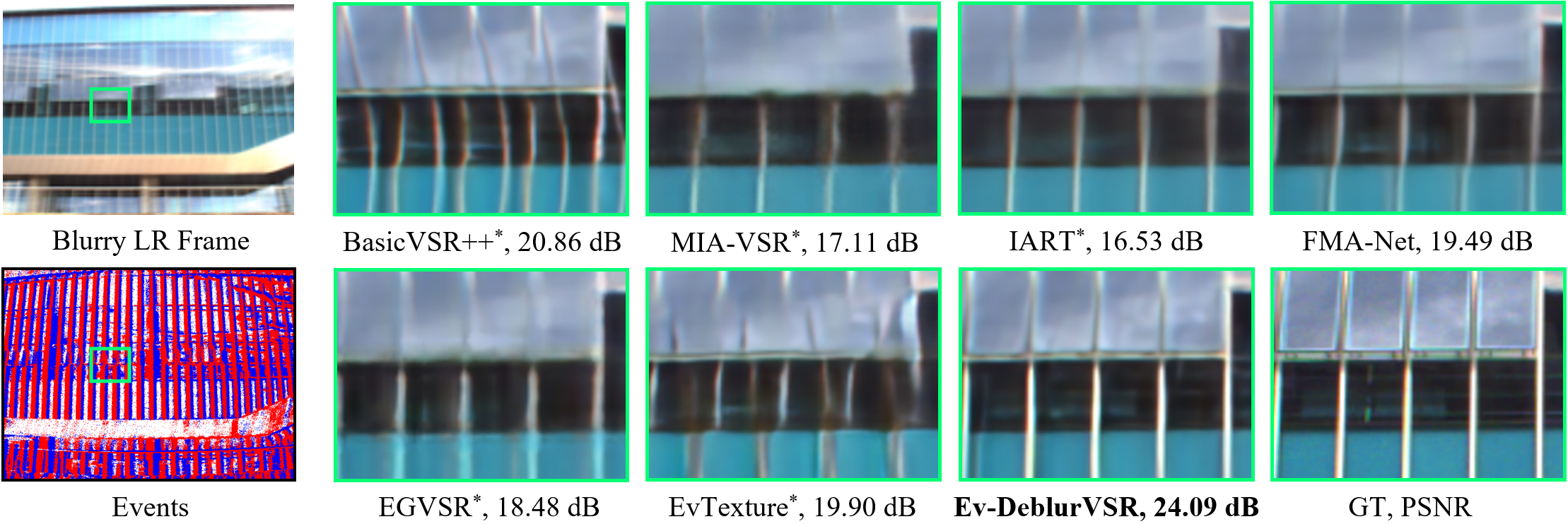}
    \caption{Qualitative comparison on NCER. Our method can restore blurred and distorted window lines to sharp ones.}
    \label{fig:fig7}
     \vspace{-1ex}
\end{figure*}

\subsection{Comparisons with State-of-the-Art Methods}

\noindent\textbf{Baselines.} We compare two types of SOTA methods: RGB-based and event-based. Each of these types is further divided into two strategies: the cascade strategy, \textit{i.e.}, deblur + VSR, and BVSR. For RGB-based VSR, we compare our method with three recent methods: BasicVSR++~\cite{chan2022basicvsr++}, MIA-VSR~\cite{zhou2024video}, and IART~\cite{xu2024enhancing}. For event-based VSR, we compare two methods: EGVSR~\cite{lu2023learning} and EvTexture~\cite{kai2024evtexture}. Also, we include two recent video deblurring methods: DSTNet~\cite{pan2023deep} and BSSTNet~\cite{zhang2024blur}, as well as two event-based motion deblurring methods: EFNet~\cite{sun2022event} and REFID~\cite{sun2023event}. Additionally, we compare with two BVSR methods: FMA-Net~\cite{youk2024fma} and eSL-Net++~\cite{yu2023learning}. It should be noted that \textbf{we retrain all baselines using the same datasets as ours for fair comparisons.}

\noindent\textbf{Quantitative results.} Tabs.~\ref{tab:table1},~\ref{tab:table2} and~\ref{tab:table3} present the comparison results against the baselines mentioned above. From the data, it is evident that our method consistently achieves superior spatial recovery in terms of PSNR, SSIM, and LPIPS~\cite{zhang2018unreasonable}, as well as temporal consistency metrics, such as tOF~\cite{chu2020learning} and TCC~\cite{chi2020all}. Notably, our method significantly improves over the recent BVSR method FMA-Net, surpassing it by \textbf{3.27} dB, \textbf{2.88} dB, and \textbf{2.59} dB on the GoPro, BSD, and NCER datasets. Additionally, our Ev-DeblurVSR has fewer parameters, requires only 33.67\% of the FLOPs, and is \textbf{7.28$\times$} faster than FMA-Net. Furthermore, our method makes better use of event data than other event-based VSR methods. In most cases, EGVSR and EvTexture do not perform better than the image-only method BasicVSR++. However, our method significantly outperforms BasicVSR++ by at least \textbf{1.55} dB across all three datasets.

\begin{table}[t!]
	\centering
	\resizebox{0.95\columnwidth}{!}{  
		\begin{tabular}{llcccc}
			\toprule
			\multicolumn{2}{c}{\multirow{2}[1]{*}{Method}} & \multicolumn{3}{c}{GoPro} & \multirow{2}[1]{*}{\makecell{\#Params \\ (M)}} \\ 
			& & PSNR$\uparrow$ & SSIM$\uparrow$ & tOF$\downarrow$  & \\
			\midrule
			\multirow{2}[1]{*}{Events} & (a) w/o inter- & 31.32 & 0.9072 & 1.67 & 7.94 \\
			& (b) w/o intra- & 31.51 & 0.9180 & 1.54 & 8.03 \\
			\midrule
			\multirow{4}[1]{*}{RFD} & (c) w/o CAB & 31.55 & 0.9176 & 1.53 & 8.25 \\
			& (d) w/o CM & 31.36 & 0.9162 & 1.58 & 8.25 \\
			& (e) w/o \textit{i$\to$e} & 31.81 & 0.9226 & 1.50 & 8.27 \\
			& (f) \textit{e$\to$i}, \textit{i$\to$e} & 32.23 & 0.9269 & 1.48 & 8.28 \\
			\midrule
			\multirow{2}[1]{*}{HDA} & (g) w/o EGA & 31.72 & 0.9163 & 1.60 & 7.98 \\
			& (h) w/o FGA & 31.53 & 0.9082 & 1.62 & 6.55 \\
			\midrule
			\multirow{2}[1]{*}{Loss} & (i) w/o $\mathcal{L}_r$ & 32.37 & 0.9288 & 1.47 & 8.28 \\
			& (j) w/o $\mathcal{L}_e$ & 32.21 & 0.9268 & 1.49 & 8.28 \\
			\midrule
			\multicolumn{2}{c}{(k) Ours} & \textbf{32.51} & \textbf{0.9314} & \textbf{1.43} & 8.28 \\
			\bottomrule
		\end{tabular}
	}
	\caption{Ablation studies of the components.} 
	\label{tab:table_abla}
  \vspace{-2ex}
\end{table}

\noindent\textbf{Qualitative results.} We also show visual comparisons in Figs.~\ref{fig:fig6} and~\ref{fig:fig7}. It is evident that our method can effectively restore clear road signs and window lines, producing sharp, well-defined edges. In contrast, other methods fail to recover fine details, resulting in blurry artifacts and indistinct boundaries. This highlights the superiority of our approach in handling blurry inputs and recovering high-quality HR frames. 

\subsection{Ablation Study}
\label{sub_abla}

\noindent\textbf{Event utilization.} Tab.~\ref{tab:table_abla}(a-b, k) shows that using only intra-frame events for both feature deblurring and alignment results in a 1.19 dB drop. This is because the timestamps of intra-frame events are not well-aligned with the nearby frames. Similarly, using only inter-frame events also causes a performance drop. Our method, which combines both intra-frame and inter-frame events, better meets the needs of BVSR, leading to a significant improvement.

\begin{figure}[t!]
	\centering
	\includegraphics[width=\columnwidth]{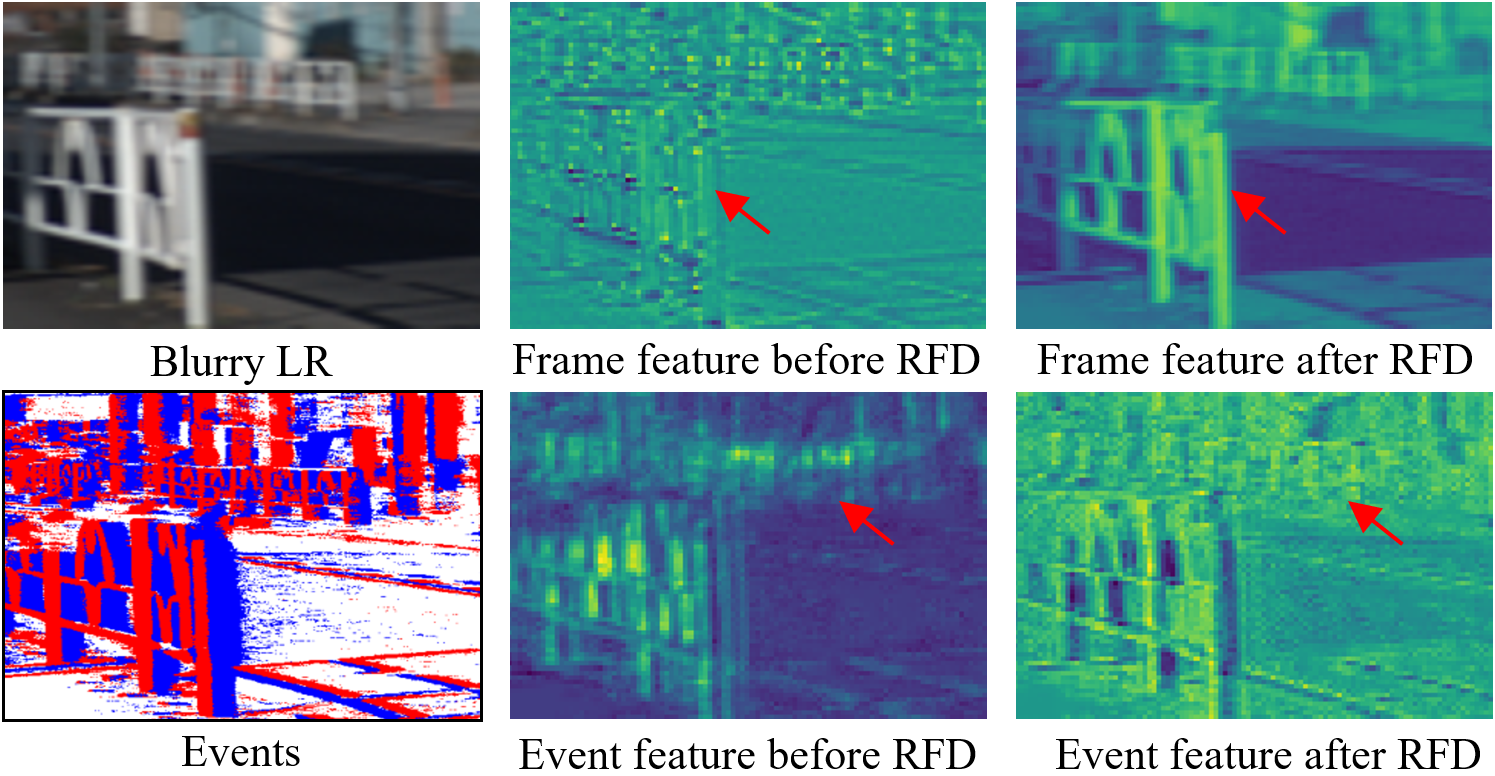}
	\caption{Analysis of the RFD module.}
	\label{fig:fig8}
 \vspace{-1ex}
\end{figure}

\begin{figure}[t!]
    \includegraphics[width=\columnwidth]{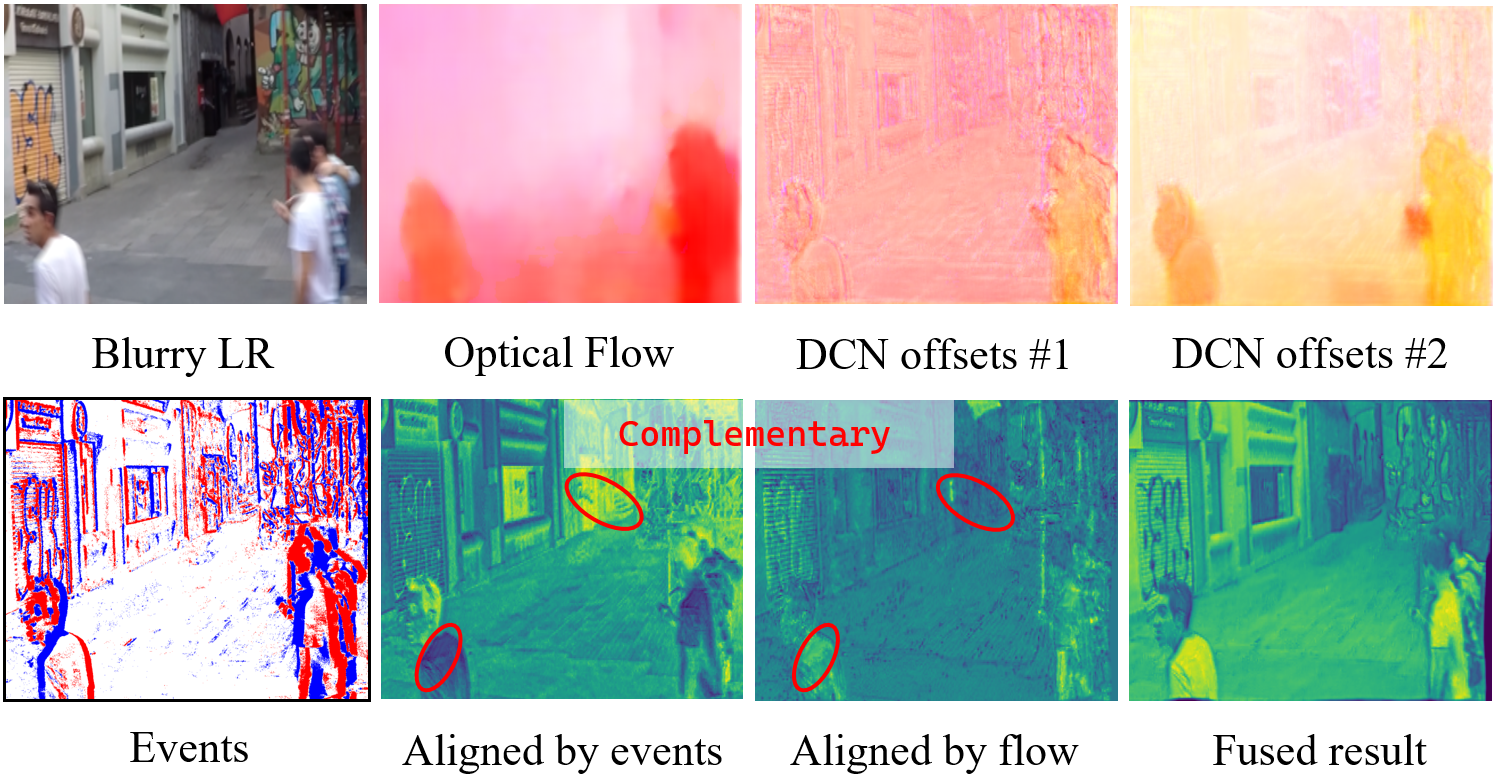}
    \caption{Analysis of the HDA module.}
    \label{fig:fig9}
     \vspace{-2ex}
\end{figure}

\noindent\textbf{The RFD module.} Tab.~\ref{tab:table_abla}(c-f, k) shows the importance of each component in our RFD module. Removing the CAB, which captures global features from event-image modalities, leads to a 0.96 dB drop. Cross-modal (CM) interaction is also critical, as its removal causes a 1.15 dB drop. In Tab.~\ref{tab:table_abla}(e-f), \textit{i$\to$e} refers to refining event features with image features, while \textit{e$\to$i }represents using event features to deblur image features. While \textit{e$\to$i} is a standard process in event-based motion deblurring, \textit{i$\to$e} is rarely explored. Excluding \textit{i$\to$e} results in a 0.70 dB drop. Our method employs a sequential \textit{i$\to$e} followed by \textit{e$\to$i}, which outperforms reversing the order, where performance drops by 0.28 dB. 

Fig.~\ref{fig:fig8} illustrates the deblurring process: in blurry areas such as railings, the RFD sharpens frame features and enriches event features with contextual scene information.

\noindent\textbf{The HDA module.} 
Tab.~\ref{tab:table_abla}(g-h, k) shows that our full model, which combines EGA and FGA alignment methods, achieves significant improvements. Fig.~\ref{fig:fig9} visualizes the learned motion vectors and aligned features. It demonstrates that DCN offsets, similar to optical flow, capture moving objects but provide more diversity. Moreover, event-aligned features can capture background areas affected by camera movement, where flow-aligned features may fail. In contrast, flow-aligned features enhance details in regions with significant motion. These two alignment methods complement each other, and the fused features exhibit sharp edges and detailed scene representations, validating the effectiveness of our hybrid alignment approach.

\begin{figure}[t!]
    \includegraphics[width=\columnwidth]{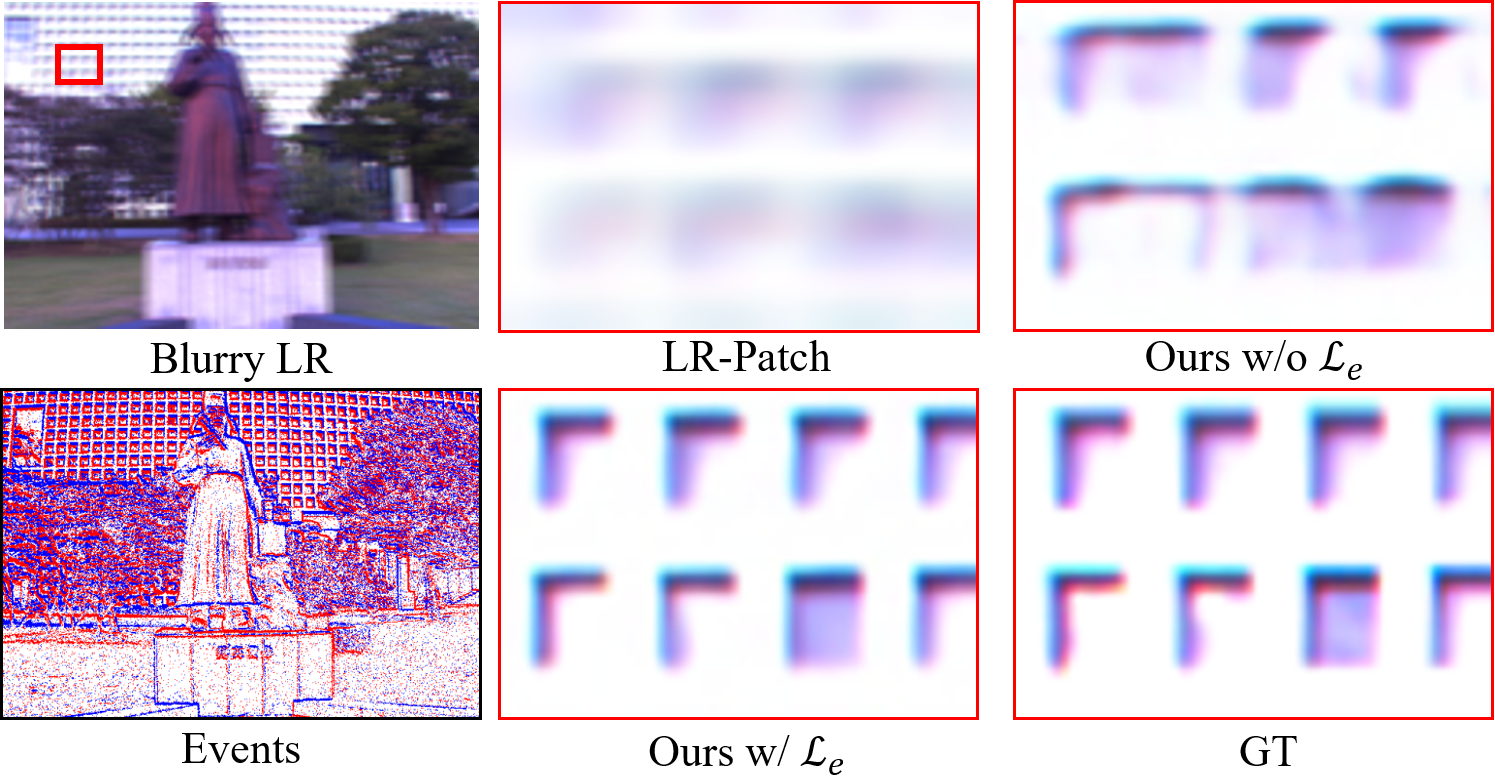}
    \caption{Analysis of the edge-enhanced loss.}
    \label{fig:fig10}
    \vspace{-1ex}
\end{figure}

\begin{table}[t!]
	\centering
	\resizebox{0.98\columnwidth}{!}{  
		\begin{tabular}{cccccc}
			\toprule
			Metrics & MIA-VSR & IART & FMA-Net & EvTexture & \textbf{Ours} \\ 
			\midrule
			PSNR$\uparrow$ & 34.14 & 33.95 & 33.00 & 34.55 & \textbf{34.99} \\
			SSIM$\uparrow$ & 0.9449 & 0.9430 & 0.9315 & 0.9491 & \textbf{0.9534}  \\
			LPIPS$\downarrow$ & 0.1695 & 0.1719 & 0.1826 & 0.1642 & \textbf{0.1551}  \\
			tOF$\downarrow$\scalebox{0.5}{$ \times  10$}  & 6.71  & 6.86 &7.22 & 6.37 & \textbf{5.98}  \\
			TCC$\uparrow$\scalebox{0.5}{$ \times  10$}  &  6.03 & 5.99 & 5.79 & 6.14 & \textbf{6.26}  \\
			\bottomrule
		\end{tabular}
	}
	\caption{Comparisons of sharp VSR methods on GoPro.} 
	\label{tab:table_sharp}
 \vspace{-2ex}
\end{table}

\noindent\textbf{Edge-enhanced loss.} Tab.~\ref{tab:table_abla}(i-j, k) demonstrates the effectiveness of our edge-enhanced loss. Fig.~\ref{fig:fig10} shows that the model trained with $\mathcal{L}_e$ more accurately restores windows, eliminating blur and producing sharper edges.

\noindent\textbf{Performance on sharp videos.} Although our primary focus is on blurry inputs, we compare our method with several recent SOTA VSR methods on sharp inputs using the GoPro dataset. As shown in Tab.~\ref{tab:table_sharp}, our method \textit{consistently} achieves the best performance on sharp inputs, both in spatial recovery and temporal consistency, demonstrating the effectiveness and versatility of our approach.

\paragraph{Limitation.} In our setting, we assume that the frame exposure time is known and fixed. However, in real-world scenarios, especially when auto-exposure is enabled, the exposure time can vary dynamically depending on the lighting conditions, making it unknown~\cite{kim2022event}. Thus, the problem of handling BVSR under unknown exposure times remains an open and worthwhile area for further research.

\section{Conclusion}

This paper presents Ev-DeblurVSR, a novel event-enhanced network for BVSR that leverages high-temporal-resolution and high-frequency event signals. To effectively fuse frame and event information for BVSR, we categorize events into intra-frame and inter-frame types. The RFD module is then introduced, utilizing intra-frame events to deblur frame features while reciprocally enhancing event features with global scene context from frames. Additionally, we propose the HDA module, which combines the complementary motion information from inter-frame events and optical flow to improve motion estimation and temporal alignment. Extensive experiments on both synthetic and real-world datasets demonstrate the effectiveness of our Ev-DeblurVSR.

\section{Acknowledgments}

We acknowledge funding from the National Natural Science Foundation of China under Grants 62472399 and 62021001.

\appendix
\section*{Appendix}

\section{More Visual Results}

\begin{figure*}[htbp]
	\centering
	\includegraphics[width=0.9\textwidth]{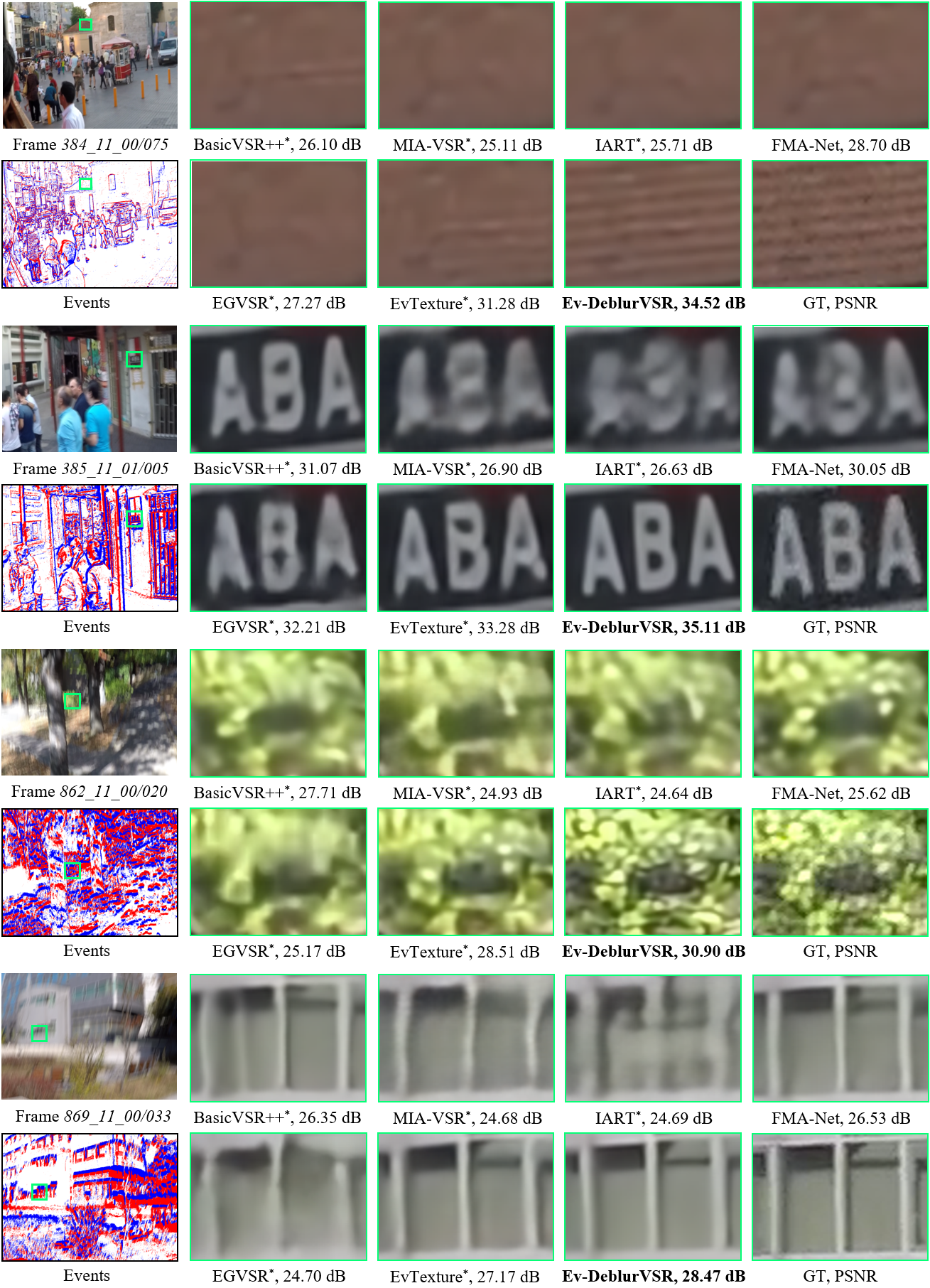}
	\caption{Qualitative comparison on GoPro~\cite{nah2017deep} for 4$\times$ BVSR. \textbf{Zoomed in for best view.}}
	\label{fig:figA1}
\end{figure*}

\begin{figure*}[htbp]
	\centering
	\includegraphics[width=0.9\textwidth]{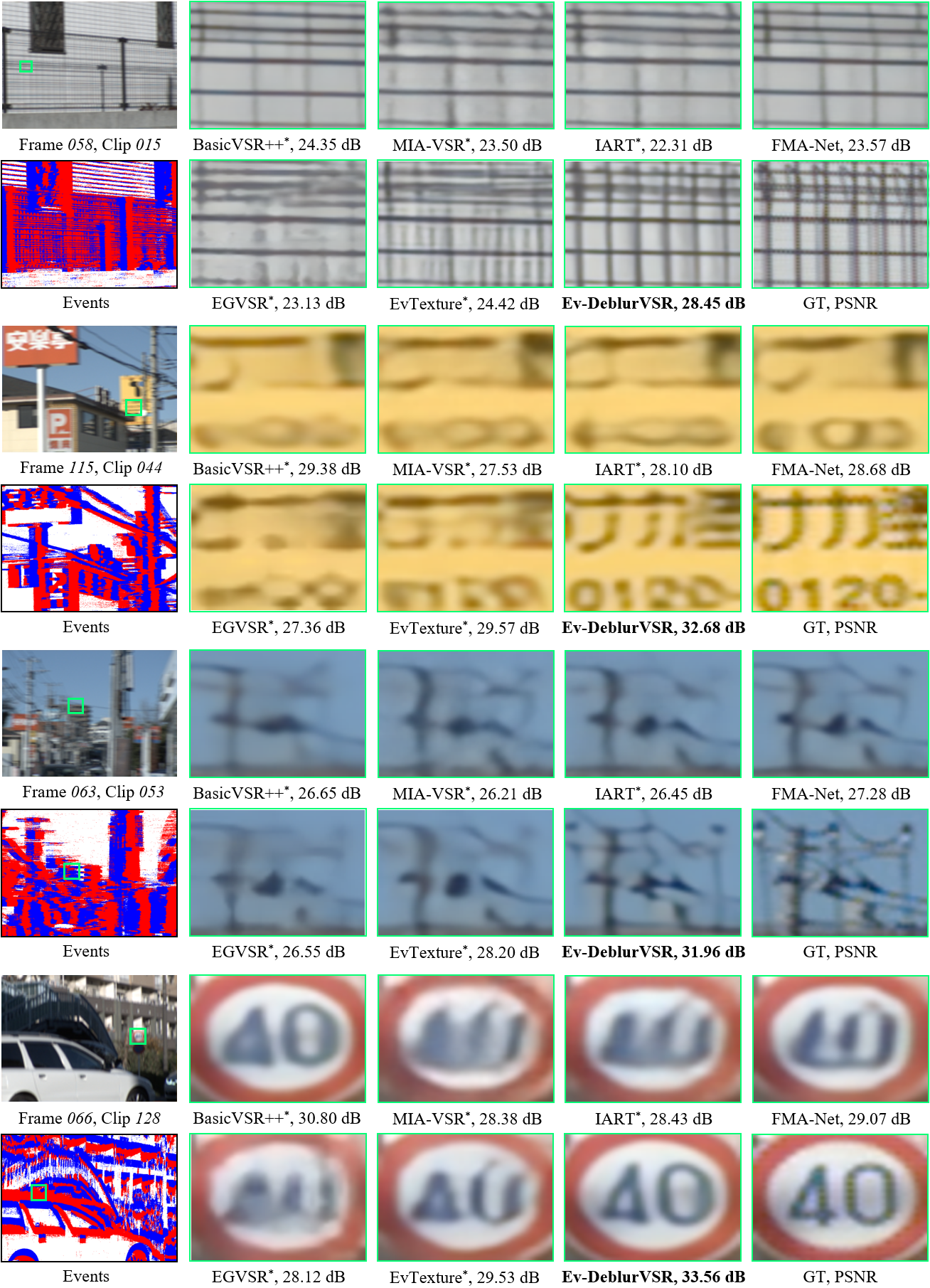}
	\caption{Qualitative comparison on BSD~\cite{zhong2020efficient} for 4$\times$ BVSR. \textbf{Zoomed in for best view.}}
	\label{fig:figA2}
\end{figure*}

\begin{figure*}[htbp]
	\centering
	\includegraphics[width=0.9\textwidth]{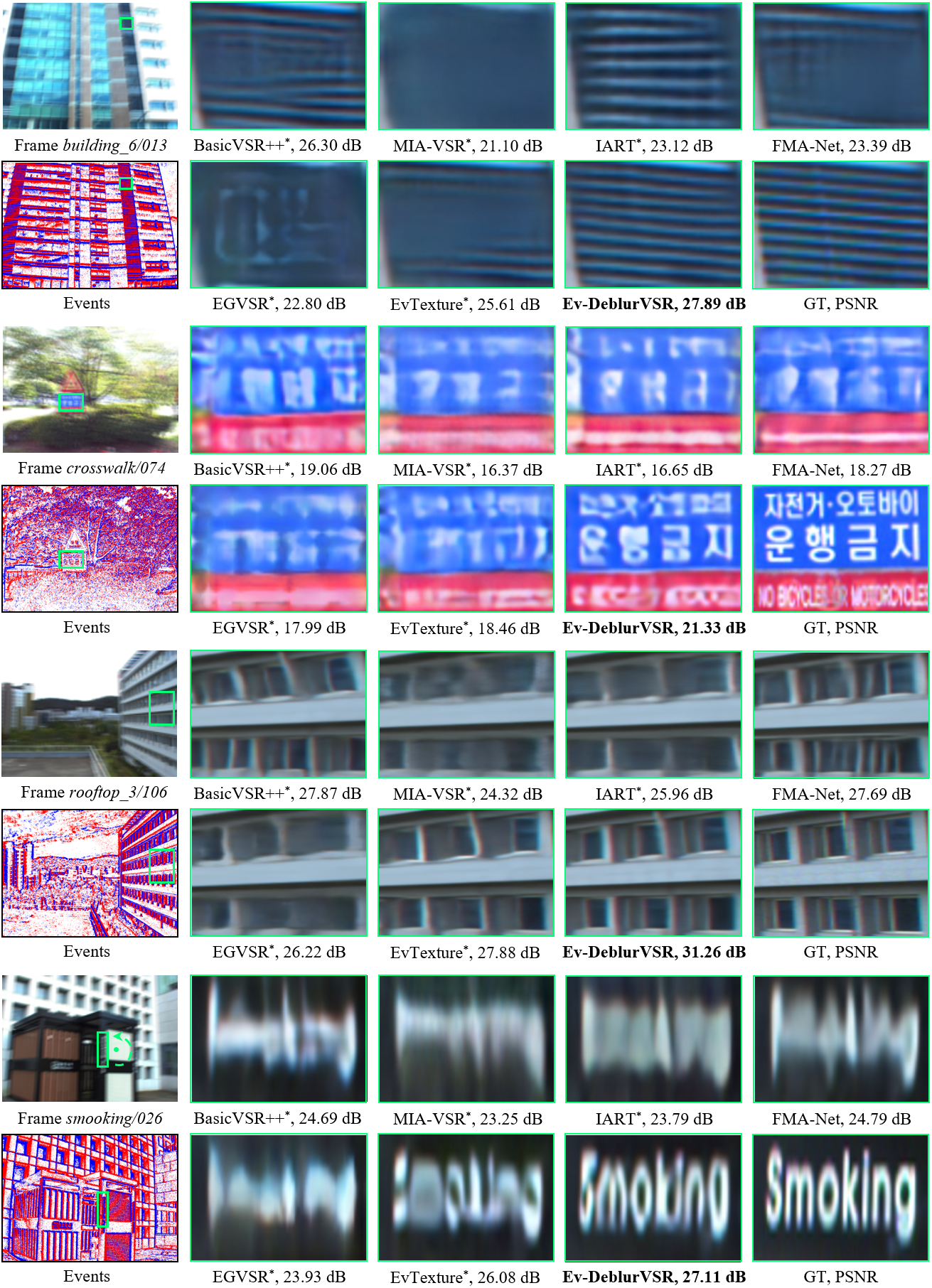}
	\caption{Qualitative comparison on NCER~\cite{cho2023non} for 4$\times$ BVSR. \textbf{Zoomed in for best view.}}
	\label{fig:figA3}
\end{figure*}

In this section, we provide additional visual comparisons on GoPro, BSD, and NCER datasets. The results are shown in Figs.~\ref{fig:figA1},~\ref{fig:figA2}, and~\ref{fig:figA3}, respectively. These results demonstrate that our Ev-DeblurVSR successfully restores complex scenes, including fine texture details on license plates, tree bark, traffic signs, and building windows, with sharp edges and minimal jitter compared to other methods.

\bibliography{aaai25}

\end{document}